# High Throughput Virtual Screening with Data Level Parallelism in Multi-core Processors

Upul Senanayake, Rahal Prabuddha and Roshan Ragel
Department of Computer Engineering
Faculty of Engineering, University of Peradeniya
Peradeniya 20400 Sri Lanka

*Abstract*— Improving the throughput of molecular docking, a computationally intensive phase of the virtual screening process, is a highly sought area of research since it has a significant weight in the drug designing process. With such improvements, the world might find cures for incurable diseases like HIV disease and Cancer sooner. Our approach presented in this paper is to utilize a multi-core environment to introduce Data Level Parallelism (DLP) to the Autodock Vina software, which is a widely used for molecular docking software. Autodock Vina already exploits Instruction Level Parallelism (ILP) in multi-core environments and therefore optimized for such environments. However, with the results we have obtained, it can be clearly seen that our approach has enhanced the throughput of the already optimized software by more than six times. This will dramatically reduce the time consumed for the lead identification phase in drug designing along with the shift in the processor technology from multi-core to many-core of the current era. Therefore, we believe that the contribution of this project will effectively make it possible to expand the number of small molecules docked against a drug target and improving the chances to design drugs for incurable diseases.

*Keywords – Virtual Screening, Drug Designing, Molecular Docking, Autodock Vina, Multi-core Processors*

## I. INTRODUCTION

The conventional approach to drug designing is an exhaustive process. It mainly concentrates on performing actual experiments on each of many candidates of molecular compounds to identify the optimal set of candidates and then further carries out experiments to find out the best binding compound to be further modified to become a drug [1]. This is undoubtedly a cost provocative way to design drugs because the researchers have to perform the lead identification process manually at the laboratories and is similar to a brute force search. Fortunately, with the development of computational biotechnology, researchers have come up with Computer Aided Drug Designing (aka CADD) to replace the initial exhaustive process of narrowing down the molecular compound candidates to the optimal set. The method we are concentrating on in our research is the molecular docking phase of a virtual screening process, which dramatically reduces the size of the ligand database (the search space for docking) from $10^{13}$ to few thousands of ligands [2].

In simple terms, the docking process is where we take a receptor protein or enzyme that causes a certain disease and match it against a ligand that could bind with the receptor in order to prevent the activities of the receptor. Virtual screening process is the repeated application of docking process using millions of compounds to narrow down the target ligand set. It is a highly computational intensive process, which is typically run on distributed grid environments for faster processing.

The idea, which drives us in optimizing such a process, is to address the issue of overhead costs in distributed grid processing with the use of parallel processing in a multi-core environment (which are going to be compatible with the many-core architecture of the future). We intend to make the comparison and identify the best option. In this specific instant, we concentrate on an infrastructure of a 32-core processor with multi-threading which effectively adds up to 64 effective threads at a time. Since Autodock Vina [3] has introduced instruction level parallelism for a multi-core environment, we will compare the results obtained using that and the data level parallelism we are trying to introduce into the picture for the same tool.

We decided to choose our protein receptor as a protease of Human Immunodeficiency Virus (HIV) [4]. There is an ongoing virtual screening process for this receptor identified as Fight Aids @ Home [5] and we intend to contribute to this project by identifying a method to utilize the multiple cores found in modern computers. We extracted the receptor 2BPW from the Protein Data Bank [6]. This receptor is more probable to represent the central tendency towards true-positives among many of HIV proteases. The ligand database was obtained from ZINC [7], which is a free database of $10^{13}$ small molecules. To be exact, for the comparison we used the first 5000 ligands from the *ChemBridge* library [8].

Following is how the rest of the paper is organized: The next section of this paper will introduce details of the virtual screening process that dramatically reduces the eligible set of ligands. Section III will be on related work that has been carried out. Section IV will demonstrate our approach to DLP in accelerating the throughput. Section V will present the results obtained from the implementation and finally in Section VI we will conclude the paper.

## II. THE VIRTUAL SCREENING PROCESS

We are describing the processes and methodologies that are used to implement the virtual screening process in order to complete the research successfully. Our final design will be directly based on the hypothesis we form in this section.

### A. Receptor Preparation

Protein receptor preparation is perhaps the most important part of this project because the whole process depends on this as a primary input. The receptor we have identified and are going to use is 2BPW receptor from Protein Data Bank, which has the structure as shown in Figure 1 [9].

The ligand included in the receptor was removed in order to prepare it for docking. We also removed the water molecules and added non-polar hydrogen atoms to the receptor using *MGLTools v1.5.4* [10]. We identified the binding site and set the grid parameters accordingly, minimizing the search space the docking algorithm needed to explore. The grid parameters define the search space which the software will use for the docking process and we have to ensure that the binding sites of the receptor are included in that.

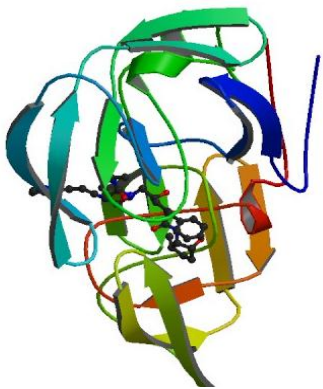

Figure 1.  Protein Receptor 2BPW

### B. Ligand Database Preparation

Since this was a control test, we considered only the first 5000 ligands from the *ChemBridge* Library ligand set. They were already prepared and were available in *PDBQT* format [21] (the ligands as well as the receptor should be in the *PDBQT* format in order to be used with Autodock Vina), which was an added advantage. Since this was a random ligand set, we consider it to be ideal to make a comparison between generic Vina that is enabled with instruction level parallelism and the data level parallelism we introduce to Vina.

### C. Docking

This is the main stage of the whole virtual screening process. So far we have prepared the receptor and ligands in such a way to facilitate the docking process. As was identified before, docking is the computationally intensive process that takes up more than 80% of the computational time. It is better understood by looking at the following illustration from Wikimedia [11]. Only one of the orientations is illustrated here although there may be several other viable orientations. We use the binding energy to pinpoint the best conformation among those identified by Autodock.

Molecular docking is similar to the *lock and key* problem. In that, we have to identify the correct orientation of the key that can be used to unlock the lock.

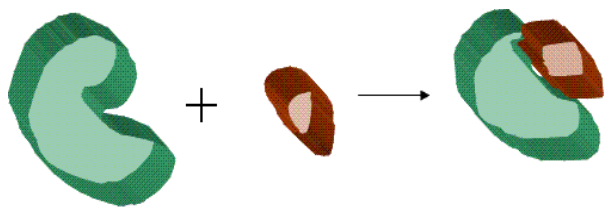

Figure 2.  Molecular Docking

We can think of the orientation as the key size, which direction the key should be turned, etc. In our context, the receptor can be thought as the lock while the ligands are the keys. Molecular docking is an optimization problem where we have to identify the best-fit orientation of the ligand that would bind to a certain receptor.

### D. Docking Mechanisms

Docking depends on two things: namely the search algorithms and the scoring functions. The search algorithms search for all possible orientations of the ligand to be bound with the receptor and it is an exhaustive search problem.

TABLE I.     DOCKING MECHANISM USED IN SOFTWARE

| Force Field Based | Empirical | Knowledge Based |
|---|---|---|
| D-Score | LUDI | PMF |
| G-Score | F-Score | DrugScore |
| GOLD | ChemScore | SMoG |
| AutoDock | SCORE | |
| Dock | Fresno | |

The energy values of protein-ligand bindings are calculated using scoring functions, which play a major role in the docking process. If the energy values are low, negative in particular, that indicates a high stability binding. There are various types of scoring functions implemented in docking programs as shown in Table I [12].

As tabulated in Table I, the scoring functions can be classified into three groups: one, force field based scoring, two, empirical ones and three, knowledge based scoring.

### E. High Throughput Virtual Screening (HTVS)

HTVS refers to the docking process that is applied in multiple times against a large number of ligands (possibly millions and growing) either sequentially or in parallel. It is more like trying to find a needle in a large haystack where some hay may resemble the characteristics of the needle that might fool you, which makes it an exhaustive search problem because until you have finished searching the whole haystack, you cannot rationally verify that you have found the optimal match.

Virtual screening was initially performed as experiments in laboratories which took ages to come up with a result and owing to the rapid development of computational biology, HTVS has now become a very important part in drug designing process, specifically in lead identification and lead optimization phases. The lead identification is the process where a feasible set of ligands are identified for a specific receptor and with the lead optimization phase, further chemical modifications are done to the ligands with the lowest binding affinity.

### F. Visualizing Results

Although visualizing results is not a main part of this comparison, it is a main part in the whole virtual screening process. The output of Vina consists of a set of models and those models have Energy Values, RMSD (root mean square deviation) values and Hydrogen Positions each. The set of models are the orientations that Vina has predicted to be bound to the receptor. We can either ask Vina to separately output those models or to integrate them to a single file and output the file. The method we used was to output the models into a single file called *out.pdbqt*.

RMSD values are calculated relative to the best mode and use only movable heavy atoms. Two variants of RMSD matrices are provided differing how the atoms are matched in the distance calculation. Due to the degrees of freedom that only move Hydrogen atoms are degenerated. Thus in the output, some Hydrogen atoms are expected to be positioned randomly but with consistent to the covalent structure. These are what are known as Hydrogen Positions. All of these characteristics are embedded into a single file and without expertise knowledge (*out.pdbqt*), it is not human readable. The recommended choice is to use PyMol Viewer [13] to visualize the results of the docking process.

### III. RELATED WORK

We have explored major virtual screening projects on similar contexts. Such projects are almost always implemented on top of a grid.

### A. WISDOM-I

WISDOM-I is related to coming up with a drug for Malaria. Its main goal is to boost research and development on neglected diseases by fostering the use of open source information technology for drug discovery as mentioned in their website [14]. They have focused on High Throughput Virtual Screening process against Malaria in 2005 and successfully completed the docking of 42 million compounds. It is mentioned that this had a cost of 80 CPU years.

### B. Virtual Screening against Influenza a Neuraminidase

This was also carried out by the same group (the group worked on *WISDOM-I*) inspired by the success they had on HTVS against Malaria. This was carried out in 2006 and it affirmed their inspiration and trust imposed on HTVS in the drug designing process. The results obtained by both experiments have been processed and drugs are being developed as we speak [15].

### C. Virtual Screening against SARS Inhibitors

Severe Acute Respiratory Syndrome (SARS) is a highly contagious upper respiratory disease that was deemed epidemic in 2002. The virtual screening process in this subject has been carried out to find the interaction patterns that should be useful in drug designing. It was run against a small set of ligands and the method it used was a combined ligand and structure based virtual screening process [16]. Since this was directed towards identifying patterns rather than identifying a drug target, the ligand database was much smaller and the search was thus carried out in a unified manner without the need of much computational resources.

A noticeable fact was that, we seldom found studies on optimizing HTVS in a multi-core environment; particularly utilizing DLP in a multi-core environment. Thus we intend to fill that vacuum with our research.

### D. Comparison

The main difference in what has been carried out in WISDOM-I and what we are proposing to carry out is that the system we are proposing will work on top of a multi-core processor environment with hyper threading while WISDOM-I concentrate on performing that operation on a distributed grid environment [17] which involves overhead costs in maintaining the nodes of the environment which we try to reduce and thus enhance the process with a time gain. Besides that, the initial process would be the same, so the advantage is that our framework and the distributed framework can easily be interchanged (or even integrated for tapping the benefits of both) if the need arises.

## IV. DATA LEVEL PARALLELISM IN VINA

We were familiar with the software Autodock Vina, which has already introduced instruction level parallelism into the picture. Thus we have decided to introduce data level parallelism to Autodock Vina and compare the results in order to identify the best alternative configuration.

### A. Proposed Architecture

We decided to consider Vina as a *black box* and continue to implement the data level parallelism around it. We looked at the possibility of a shared memory in case there are dependencies; however, since the unit of execution we considered was an instance of Vina and for that, the input was only the receptor and a ligand we were able to surpass the dependency issues.

We understood that the best way to implement data level parallelism is to provide an abstract layer that handles the number of jobs (threads/instances) it executes at any given time. We can define the number of jobs and the layer would spawn a new instance of Vina till the layer has a defined number of jobs. Immediately after a job finishes, the layer spawns another instance of Vina and so forth. The diagram in Figure 3 illustrates the architecture.

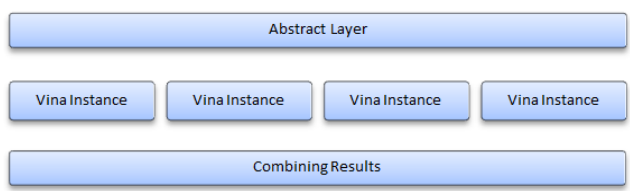

Figure 3. Illustration of the Abstract Layer

It should be noted that wherever the number of CPU configuration is used denotes the number of CPU configuration inherent to Vina and the number of Vina instances denotes the number of jobs we would introduce with the data level parallelism implementation.

### B. Dry Run of Data Level Parallelism

Since Vina was already optimized to work in a multi-core environment, we had to explore whether we can make some further enhancement by introducing data level parallelism into it. In the dry run, we came up with a simple shell script with Job Control facilities to substantiate our hypothesis. We used a set of 16 ligands to be tested using Vina and the script we came up with. Repeated experiments were carried out to obtain an average value for the elapsed time to compensate the possibility of CPU cycles been utilized for other system functions.

Hypothesis: *Data level parallelization can introduce more throughput enhancement than instruction level parallelism.*

The shell script we used was inspired from the work done by Jure Varlec on *prll* [18] which acts as a wrapper to run jobs in parallel using a shell. It enables us to define how many concurrent jobs we want. It uses the built-in *jobs* command of shell and depending on the number of jobs we defined; if it has sufficient jobs running, the script sleeps. Every .33s it checks this condition and if it finds an idle slot, another job is initiated. It should be noted that, this was a very low-level data parallelism since it is a mere implementation of running several jobs together without a job scheduling mechanism. Our reasoning behind using such a technique is that if we can introduce a speed enhancement with a parallelism of this calibre, we can definitely work towards further improving it.

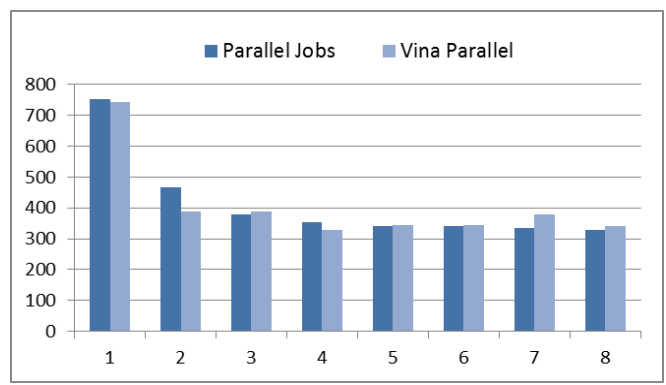

Figure 4. Variation of Elapsed Time (s) for Various Configurations

The results were indeed self-explanatory. We identified that the data level parallelism introduced in 8 Vina instances configuration yields the best results among all. We ran this test on Intel Core i3 Processor with 64-bit operating System (Ubuntu 11.10) and 4GB of RAM. As it can be seen from the graph in Figure 4, the equivalent configuration in Vina also yields almost the same results, but we were positive that given a larger number of ligands to be screened, the data level parallelism would yield a better throughput enhancement.

## C. Data Level Parallelism in Python

We have decided to use *python* [19] for this research because it was easy to use and had a variety of libraries that we can use in order to achieve our task. We were exploring symmetric multi-processing libraries in *python* and the *joblib* scheduling library [20] was identified as the best choice for our cause.

## D. Joblib Library

As mentioned in the vision of creating *joblib* library, it provides tools that achieve better performance and reproducibility when working with long running jobs. We have the advantage of not having to modify our code because *joblib* only introduces job pipelining. The reason for this to enhance speed compared to instruction level parallelism would be the differences in ligands. *Joblib* facilitates on-demand computing and transparent parallelization, which is important for our observation since we need to have a controlled environment. Further, since *joblib* doesn't have any dependencies other than *python*, it is convenient to use in a remote machine.

## E. Implementation

The configurations we took into consideration are listed in Table II. As we mentioned, there was the need to make a thorough comparison in order to come up with a conclusion because Vina also can exploit the power of multiple cores to its ILP. Thus we decided the configurations in Table II would be ideal. We have also mentioned the average time taken to execute these jobs from the 5 iterations we ran. Since the average is taken, we are assuming the full CPU cycle utilization was available for this process.

## V. RESULTS AND DISCUSSION

The docking experiments reported in Table II were carried out against a set of 5000 ligands from the ChemBridge data library, which can be considered as an accurate cross section of ligands. As it can be clearly seen, the average time taken using the configurations 32 | 4 (32 jobs and 4 cores in Vina) is at the minimum while running Vina with a single core yields the maximum time. The latter configuration is self-explanatory since the single core operation should definitely take more time than any of the multi-core operations for the same test set. From that point onwards, the average elapsed time has gradually reduced under multi-core utilization. We will discuss about the differences between these results sets later in this section. The visualization of the data set represented in Table II is shown in Figure 5.

The graph clearly shows the differences between the respective configurations. It is worthwhile to note that the vertical axis is on logarithmic scale because there are extremely high values compared to other values.

TABLE II. AVERAGE ELAPSED TIME FOR DIFFERENT CONFIGURATIONS

| # of Jobs | # of CPUs in Vina | Average Time Taken [s] | Standard Deviation % |
|---|---|---|---|
| 64 | 1 | 12554 | 0.11 |
| 64 | 2 | 12541 | 0.07 |
| 64 | 4 | 12622 | 0.24 |
| 32 | 1 | 15273 | 0.45 |
| 32 | 2 | 12652 | 0.07 |
| 32 | 4 | 12501 | 0.13 |
| 16 | 1 | 26915 | 0.08 |
| 16 | 2 | 15464 | 0.25 |
| 16 | 4 | 12909 | 0.05 |
| 8 | 1 | 53584 | 0.07 |
| 8 | 2 | 27767 | 0.10 |
| 8 | 4 | 16743 | 0.18 |
| N/A | 1 | 429636 | 0.04 |
| N/A | 2 | 228081 | 0.21 |
| N/A | 4 | 122309 | 0.69 |
| N/A | 8 | 76568 | 0.06 |
| N/A | 64 | 79522 | 0.12 |

We explored the possibility of introducing *data level parallelism* to Vina. We performed a dry run to verify the fact that we can actually introduce an enhancement and after that, we started the implementation. We have identified *python* as our programming language and decided to use *joblib* library for pipelining the jobs. The implementation consisted of accessing ligand files in separate jobs until all the cores are occupied. The scheduling was taken care by *joblib*, thus we implemented the repetitive application of Vina in parallel and unifying the set of results obtained by the docking process.

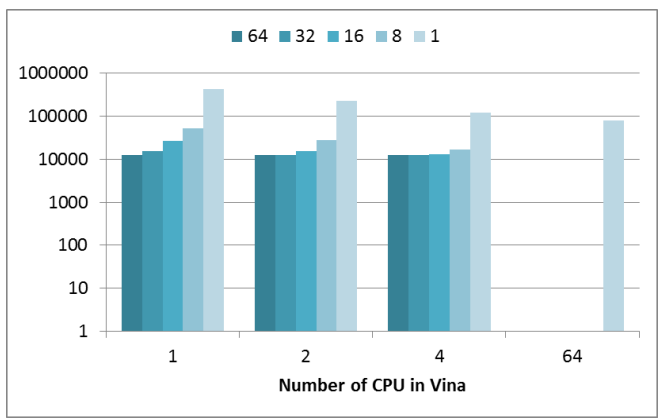

Figure 5. Comparison of Elapsed Time

As the result depicts, our implementation turned out to be a success with a considerable margin. Let us tabulate the results obtained and the speed enhancement each configuration has introduced in Table III. The second column compares the throughput enhancement with respect to a Vina instance with no parallelism whatsoever and the final column compares the throughput enhancement against the ILP that is already available in Vina. Both are in terms of multipliers. It is conclusive that data level parallelism can indeed introduce a throughput enhancement that is more than six times that of Vina with ILP. If we are to consider Vina without ILP, then the throughput enhancement is as much as 34 times.

This can be explained well if we consider the ligands separately. According to the initial dry runs we performed, a docking operation can take up to 30s to 360s depending on the ligand in question. Thus when we use data level parallelism, the smallest data unit will be a ligand and a separate Vina instance is called for each job we pipeline. However, since our scheduling makes sure that a core is not free at any given time, the full number of cores as defined from the number of jobs parameter is utilized. This addresses the different times taken for a ligand to finish the docking process.

TABLE III. PERCENTAGE ENHANCEMENT AGAINST EACH CONFIGURATION

| Configuration | Enhancement w.r.t. Single Core Environment (x) | Enhancement w.r.t. similar Multi Core Environment (x) |
|---|---|---|
| 64 \| 1 | 34.22 | 6.33 |
| 32 \| 2 | 33.95 | 6.22 |
| 16 \| 4 | 33.28 | 6.03 |

On the other hand, using Vina's built in instruction level parallelism does not guarantee the maximum utilization of CPU cycles because there is only so much of instruction level parallelism one can introduce for an operation without messing up with the dependencies. In the case of docking, sometimes, Vina needs to wait for other confirmations before processing and thus creating a dependency that needs to be met. Thus it logically follows that Vina's inbuilt parallelism will not take the full usage of the multi-core environment. We hypothesized this before the experiment and our final results have confirmed that.

VI. CONCLUSION

Drug designing for certain diseases has been one of the most compelling tasks sought by researchers. We have chosen a receptor that is linked to HIV protease which has been tested for positive drug targets actively in various high throughput systems. It is perhaps one of the critical problems that need to be solved in this century. With our approach, researchers can dramatically reduce the time taken for the HTVS process effectively making them to go through more ligands at the same duration of time. Further, since it also eliminates overhead costs incurred in similar grid infrastructures, this solution may prove to be a turning point in the history of drug designing.